\documentclass[preprint,5p,12pt]{elsarticle}
\usepackage{times}
\usepackage{epsfig}
\usepackage{graphicx}
\usepackage{amsmath}
\usepackage{amssymb}
\usepackage{fancybox}
\usepackage{alltt}
\usepackage{soul}
\usepackage{color}
\usepackage{verbatim}
\usepackage{xcolor}
\usepackage{colortbl,hhline}
\usepackage{algorithmic}
\usepackage[ruled,vlined]{algorithm2e}
\usepackage{framed}
\usepackage{pgf}
\usepackage{tikz}
\usetikzlibrary{arrows,automata}
\usetikzlibrary{positioning}

\usepackage[pagebackref=true,breaklinks=true,letterpaper=true,colorlinks,
bookmarks=false]{hyperref}

\journal{Pattern Recognition}

\begin{document}

\begin{frontmatter}
\title{A supervised clustering approach for \emph{fMRI}-based inference of
brain states}
\author[inria]{Vincent Michel}
\author[inria]{Alexandre Gramfort}
\author[inria]{Ga\"el Varoquaux}
\author[inserm]{Evelyn Eger}
\author[select,upsud]{Christine Keribin}
\author[inria]{Bertrand Thirion\corref{cor1}}

\address[inria]{Parietal team INRIA Saclay-Ile-de-France, France.}
\address[inserm]{INSERM U562, Gif/Yvette, France - CEA/DSV/I2BM/Neurospin/LCOGN}
\address[select]{ Select team INRIA Saclay-Ile-de-France, France.}
\address[upsud]{Universit\'e Paris Sud, Laboratoire de Math\'ematiques, UMR
8628, Orsay, France.}

\cortext[cor1]{Corresponding author: bertrand.thirion@inria.fr}

\begin{abstract}
We propose a method that combines signals from
many brain regions observed in functional Magnetic Resonance Imaging (fMRI)
to predict the subject's behavior during a scanning session.
Such predictions suffer from the huge number of brain
regions sampled on the voxel grid of standard fMRI data sets: the 
curse of dimensionality.
Dimensionality reduction is thus needed, but it is often performed using a
univariate feature selection procedure, that handles neither the spatial 
structure of the images, nor the multivariate nature of the signal.
By introducing a hierarchical clustering of the brain volume that
incorporates connectivity constraints, we reduce the span of the possible
spatial configurations to a single tree of nested regions tailored to the 
signal.
We then prune the tree in a supervised setting, hence the name \emph{supervised
clustering}, in order to extract 
a parcellation (division of the volume)
such that parcel-based signal averages best predict the target information.
 Dimensionality reduction is thus achieved by \emph{feature
agglomeration}, and the constructed features now provide a multi-scale
representation of the signal.
Comparisons with reference methods on both simulated and real data
show that our approach yields higher prediction accuracy than
standard voxel-based approaches.
Moreover, the method infers an explicit weighting of the regions
involved in the regression or classification task.
\end{abstract}


\end{frontmatter}


\section{Introduction}
\label{sec:intro}

Inferring behavior information or cognitive states from brain activation
images (\emph{a.k.a. inverse inference}) such as those obtained with
functional Magnetic Resonance Imaging (fMRI)
is a recent approach in neuroimaging \cite{cox2003} that
can provide more sensitive analysis than standard statistical
parametric mapping procedures \cite{kamitani2005}.
Specifically, it can be used to assess the involvement of some brain
regions in certain cognitive, motor or perceptual functions, by evaluating
the accuracy of the prediction of a behavioral variable of interest
(the \emph{target}) when the classifier is instantiated on these
brain regions.
Such an approach can be particularly well suited for the investigation of
coding principles in the brain \cite{dayan2001}. Indeed certain neuronal populations
activate
specifically when a certain perceptual or
cognitive parameter reaches a given value. Inferring the parameter
from the neuronal activity and extracting the \emph{spatial organization} of
this
coding helps to \emph{decode} the brain system.

Brain decoding requires to define a prediction function such as a
classifier that relates the image data to relevant variables.
Many methods have been tested for classification or regression of
activation images (Linear Discriminant Analysis, Support Vector
Machines, Lasso, Elastic net regression and many others), but in this
problem the major bottleneck remains the localization of predictive
regions within the brain volume (see \cite{haynes2006} for a
review).
Selection of relevant regions, \emph{a.k.a.} feature selection, is important
both to achieve accurate prediction (by
alleviating the curse of dimensionality) and understand the spatial
distribution of the informative features \cite{carroll2009}.
In particular, when the number of \emph{features} (voxels, regions) is much
larger ($\sim 10^5$) than the numbers of samples (images) ($\sim 10^2$), the
prediction method \emph{overfits} the training set, and thus does not generalize
well.
To date, the most widely used method for feature selection is
voxel-based Anova (Analysis of Variance), that evaluates each brain
voxel independently. The features that it selects can be redundant,
and are not constrained by spatial information, so that they can be spread
across all brain regions.  Such maps are difficult to interpret,
especially compared to standard brain mapping techniques such as
\emph{Statistical Parametric Maps}
\cite{friston1995c}. Constructing spatially-informed predictive
features gi\-ves access to meaningful maps (\emph{e.g.} by
constructing informative and anatomically coherent regions
\cite{cordes2002}) within the decoding framework of \emph{inverse
  inference}.

A first solution is to introduce the spatial information within a voxel-based
analysis, \emph{e.g.} by adding region-based
priors \cite{palatucci2007}, by using a spatially-informed regularization
\cite{michel2010a} or by keeping only the neighboring voxels for the predictive
model, such as in the \emph{searchlight} approach \cite{kriegeskorte2006};
however the latter approach cannot handle long-range interactions in the
information coding.

A more natural way for using the spatial information  is called \emph{feature
agglomeration}, and consists of replacing voxel-based signals by local
averages (\emph{a.k.a.}
\emph{parcels}) \cite{flandin2002,mitchell2004,fan2006,thirion2006}.
This is motivated by the fact that \emph{fMRI} signal has a strong spatial
coherence due to the spatial extension of the underlying metabolic changes and
of the neural code \cite{ugurbil2003}. There is a local redundancy of the
predictive
information.
Using these parcel-based averages of fMRI signals to fit the \emph{target}
naturally reduces the number of features (from $\sim 10^5$ voxels to $\sim
10^2$ parcels).
These parcels can be created using only spatial information, in a
purely geometrical approach \cite{kontos2004}, or using atlases
\cite{mazoyer2002,keller2009}.  In order to take into account both
spatial information and functional data, clustering approaches have
also been proposed, \emph{e.g.} spectral clustering
\cite{thirion2006}, Gaussian mixture models \cite{thyreau2006},
K-means \cite{ghebreab2008} or fuzzy clustering \cite{he2008}. The
optimal number of clusters may be hard to find
\cite{thyreau2006,filzmoser1999}, but probabilistic clustering
provides a solution \cite{tucholka2008}.
Moreover, as such spatial averages can lose the fine-grained
information, which is crucial for an accurate decoding of fMRI data
\cite{cox2003,haynes2006,haynes2005}, different resolutions of information
should be allowed \cite{golland2007}.

In this article, we present a \emph{supervised clustering} algorithm,
that \emph{considers the target to
be predicted during the clustering procedure} and yields an adaptive 
segmentation into \emph{both} large regions and
fine-grained information, and can thus be considered as \emph{multi-scale}.
The proposed approach is a generalization of \cite{michel2010} usable with 
any type of prediction functions, in both classification and regression
settings.
%
\emph{Supervised clustering} is presented in section \ref{sec:method}, and is
illustrated in section \ref{sec:simu} on simulated data.
In section \ref{sec:results}, we show on real \emph{fMRI} data sets in
regression and classification settings, that our method can recover
the discriminative pattern embedded in an image while
yielding higher prediction performance than previous approaches. Moreover, \emph{supervised clustering}
appears to be a powerful approach for the challenging generalization across
subjects (inter-subject \emph{inverse inference}).
%


\section{Methods}
\label{sec:method}


\subsubsection*{Predictive  linear model}
Let us introduce the following predictive linear model for regression settings:
\begin{equation}
\bold{y} = \bold{X}\, \bold{w} + b \enspace ,
\label{Eq:model_reg}
\end{equation}
where $\bold{y}  \in \mathbb{R}^n$ represents the behavior variable and
$(\bold{w},b)$ are the parameters to be estimated on a training set comprising
$n$ samples. A vector $\bold{w} \in \mathbb{R}^p$ can be seen as an image;
$p$ is the number of features (or voxels) and $b \in \mathbb{R}$
is called the \emph{intercept} (or \emph{bias}). The matrix $\bold{X} 
\in \mathbb{R}^{n\times p}$
is the design matrix. Each row is a $p$-dimensional sample, \emph{i.e.},
an activation map related to the observation.
In the case of classification with a linear model, we have:
\begin{equation}
\bold{y}  = \mbox{sign} (\bold{X}\, \bold{w} + b),
\label{Eq:model_classif}
\end{equation}
where $\bold{y} \in \{-1,1\}^n$ and ``$\mbox{sign}$'' denotes the sign function.
The use of the intercept is fundamental in practice as it allows the
separating hyperplane to be offseted from 0. However for the sake
of simplicity in the presentation of the method, we will from now on
consider $b$ as an added coefficient in the vector $\bold{w}$. This is done by
concatenating a column filled with 1 to the matrix
$\bold{X}$. We note $\bold{X}^j$ the signal in the $j^{th}$ voxel (feature)
$v^j$.

\subsubsection*{Parcels}

We define a \emph{parcel} $P$ as a group of connected voxels, a
\emph{parcellation} $\mathcal{P}$ being a partition of the whole set of features
in a set of \emph{parcels}:
\begin{equation}
\forall j \in [1,\ldots,p] \;,\; \exists k \in [1,\ldots,\delta] \;:\; 
v^j \in P^{k},
\end{equation}
such that
\begin{equation}
\forall (k,k)' \in [1,\ldots,\delta]^2 \; s.t. \; k\neq k', \; P^k \cap  P^{k'} = \emptyset
\end{equation}
where $\delta$ is the number of parcels and $P^{k}$ the $k^{th}$
parcel.  The \emph{parcel-based signal} $\bold{X_p}$ is the average
signal of all voxels within each parcel (other representation can be
considered, \emph{e.g.} median values of each parcel), and the
$k^{th}$ row of $\bold{X_p}$ is noted $\bold{X_p}^k$:
\begin{equation}
\bold{X_p}^k = \frac{\sum_{j |v^j \in P^k} \bold{X}^j}{p_k}
\end{equation}
where $p_k$ is the number of voxels in the parcel $P^k$.

\subsubsection*{Bayesian Ridge Regression}

We now detail \emph{Bayesian Ridge Regression (BRR)} which is the predictive
linear model used for regression in this article, and we give implementation
details on parcel-based averages $\bold{X_p}$.
\emph{BRR}  is based on the following Gaussian assumption:
\begin{equation}
p(\bold{y}|\bold{X_p},\bold{w},\alpha) =
\prod_{i=1}^{i=N} \mathcal{N}(y_i|\bold{X_{p,i}}\bold{w},\alpha^{-1})
\end{equation}
We assume that the noise $\epsilon$ is Gaussian with  precision 
$\alpha$ (inverse of
 the variance), \emph{i.e.} $p(\epsilon|\alpha)=
\mathcal{N}(0,\alpha^{-1} \bold{I_{n}})$.
For \emph{regularization} purpose, \emph{i.e.} by constraining
the values of the weights to be small, one can add a Gaussian prior
on $\bold{w}$, \emph{i.e.} $p(\bold{w}|\lambda)
= \mathcal{N}(\bold{w}|0,\lambda^{-1}\bold{I_{p}})$, that leads to:
\begin{equation}
p(\bold{w}|\bold{X_p},\bold{y},\alpha,\lambda)
\propto 
\mathcal{N}(\bold{w}|\bold{\mu},\Sigma) \;\;,\;\;
\end{equation}
where:
\begin{eqnarray}
\left\{
\begin{array}{l}
\bold{\mu} =  \alpha \Sigma \bold{X_p}^T \bold{y}\\
\bold{\Sigma} =  (\lambda \bold{I_p} + \alpha \bold{X_p}^T \bold{X_p})^{-1}
\end{array}
\right.
\label{Eq:brr_mu_sigma}
\end{eqnarray}

In order to have a full Bayesian framework and to avoid degenerate solutions,
one can add classical Gamma
priors on $\alpha \sim \Gamma(\alpha;\alpha_{1},\alpha_{2}) $ and $\lambda
\sim \Gamma(\lambda;\lambda_{1},\lambda_{2})$:
\begin{equation}
\Gamma(x;x_{1},x_{2}) = x_{2}^{x_{1}}
x^{x_{1}-1} \frac{\exp^{-xx_{2}}}{\Gamma(x_{1})}
\end{equation}
and the parameters update reads:
\begin{eqnarray}
\left\{
\begin{array}{l}
\hat{\bold{\lambda}} = \frac{\gamma + 2\lambda_1}{\bold{\mu}^T
\bold{\mu}+ 2\lambda_2}\\
\hat{\bold{\alpha}} = \frac{n - \gamma + 2\alpha_1}
{\sum_{i=1}^{i=n}(y_i -\bold{X_{p,i}}\bold{\mu})^{2}  + 2\alpha_2}  \;\;,\;\;
\end{array}
\right.
\label{Eq:brr_lambda_alpha}
\end{eqnarray}
where $ \gamma = \sum_{i=1}^{i=p}\frac{\alpha s_i}{\lambda + \alpha s_i}$,
and $s_i$ are the eigenvalues of $\bold{X_p}^T \bold{X_p}$.
In the experiments detailed in this article, we choose  $\lambda_1 = \lambda_2 =
\alpha_1 = \alpha_2 = 10^{-6}$, \emph{i.e.} weakly informative priors.

\emph{BRR} is solved using an iterative algorithm that maximizes the
\emph{log likelihood}; starting with $\alpha =
\frac{1}{\mbox{var}(\bold{y}^t)}$ and $\lambda = 1$, we iteratively
evaluate $\bold{\mu}$ and $\bold{\Sigma}$ using
Eq.~(\ref{Eq:brr_mu_sigma}), and use these values to estimate
$\gamma$, $\hat{\bold{\lambda}}$ and $\hat{\bold{\alpha}}$, using
Eq.~(\ref{Eq:brr_lambda_alpha}). The convergence of the algorithm is
monitored by the updates of $\bold{w}$, and the algorithm is stopped
if $\|\bold{w}_{s+1} - \bold{w}_{s}\|^1 < 10^{-3}$, where
$\bold{w}_{s}$ and $\bold{w}_{s+1}$ are the values of $\bold{w}$ in
two consecutive steps.

\subsection{Supervised clustering}

\begin{figure*}[h!tb]
\center \includegraphics[width = 1.\linewidth]{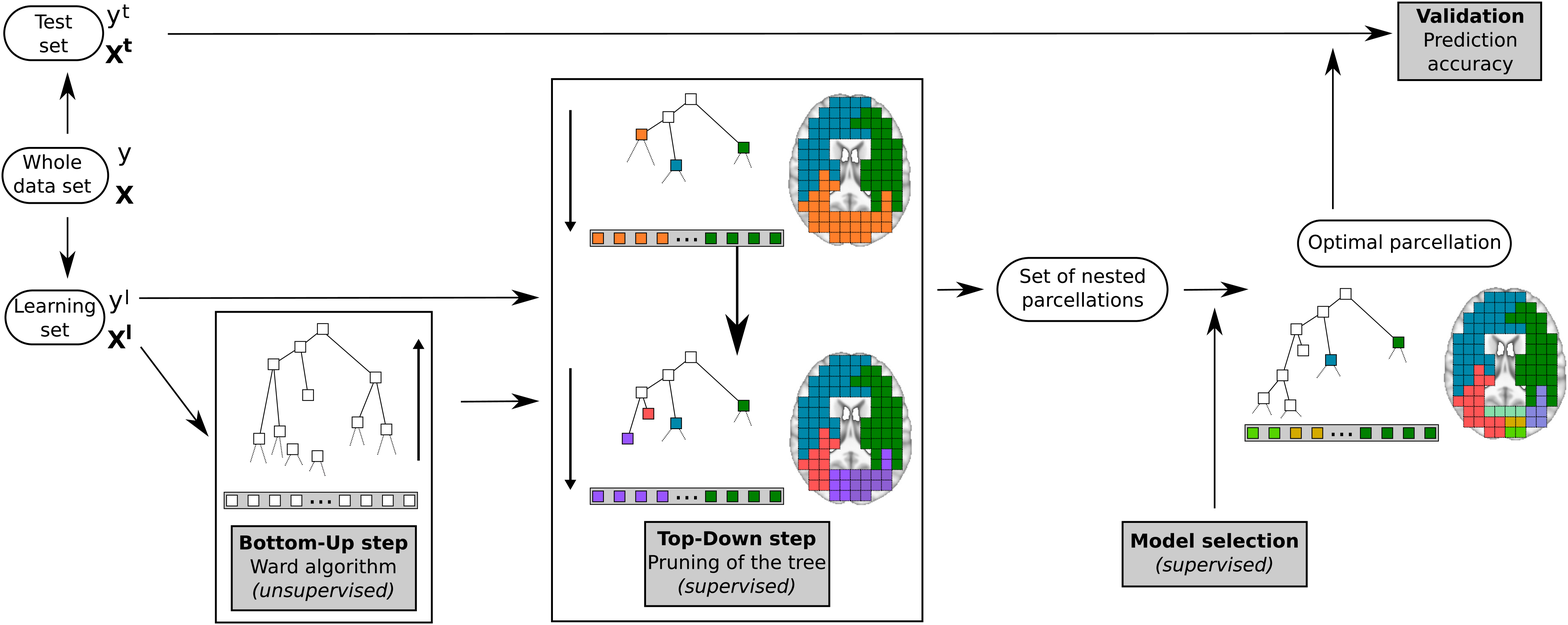}
\caption{Flowchart of the \emph{supervised clustering} approach.
\emph{Bottom-Up step (Ward clustering)} - step \ref{Step:hc}: the tree
$\mathcal{T}$ is constructed from the leaves (the voxels in the gray
box) to the unique root (\emph{i.e.} the full brain volume), following
spatial connectivity constraints.
\emph{Top-Down step (Pruning of the tree)} - step \ref{Step:pruning}: the Ward's
tree is cut recursively into smaller sub-trees, each one
corresponding to a parcellation, in order to maximize a prediction
accuracy $\zeta$.
\emph{Model selection} - step \ref{Step:sel}: given the set of nested
parcellations obtained by the pruning step, we select the optimal sub-tree
$\widehat{\mathcal{T}}$, i.e. the one that yields the optimal value for $\zeta$.
}
\label{Fig:sc_flowchart}
\end{figure*}

In this section, we detail an original contribution, called
\emph{supervised clustering}, which addresses the limitations of the
\emph{unsupervised feature agglomeration} approaches.
The flowchart of the proposed approach is given in Fig.~\ref{Fig:sc_flowchart}.

We first construct a hierarchical subdivision of the search domain using Ward
hierarchical clustering  algorithm~\cite{ward1963}.
The resulting nested parcel sets constructed from the functional
data is isomorphic to a tree. By construction, there is a one-to-one mapping
between cuts
of this tree and parcellations of the domain. Given a parcellation, the
signal can be represented by parcel-based averages, thus providing a low
dimensional representation of the data (\emph{i.e. feature agglomeration}).
The method proposed in this
contribution is a greedy approach that optimizes the cut in order to maximize
the prediction accuracy based on the parcel-based averages. By doing so, a
parcellation of the domain is 
estimated in a supervised learning setting, hence the name \emph{supervised
clustering}. We now detail the different steps of the procedure.

\subsubsection{Bottom-Up step: hierarchical clustering}
\label{Step:hc}

In the first step, we ignore the target information -- \emph{i.e.} the
behavioral variable to be predicted -- and use a \emph{hierarchical
agglomerative clustering}.
We add connectivity constraints to this algorithm (only adjacent clusters
can be merged together) so that only spatially connected \emph{clusters},
\emph{i.e.} \emph{parcels}, are created.
This approach creates a hierarchy of parcels represented as a tree $\mathcal{T}$
(or dendrogram)
\cite{johnson1967}.
As the resulting nested parcel sets is isomorphic to the tree $\mathcal{T}$, we
identify any tree cut with a given parcellation of
the domain.  The root of the tree is the unique parcel that gathers all the
voxels, the leaves being the parcels with only one voxel.
Any cut of the tree into $\delta$ sub-trees corresponds to a unique parcellation
$\mathcal{P}_\delta$, through which the
data can be
reduced to $\delta$ parcels-based averages.
Among different \emph{hierarchical agglomerative clustering}, we use the
variance-minimizing approach of Ward algorithm \cite{ward1963} in order to
ensure that \emph{parcel-based} averages provide a
fair representation of the signal within each parcel.
At each step, we merge together the two \emph{parcels} so that the resulting
parcellation minimizes the sum of squared differences within all \emph{parcels}
(\emph{inertia criterion}).

\subsubsection{Top-Down step: pruning of the tree $\mathcal{T}$}
\label{Step:pruning}

\begin{figure}
\center \includegraphics[width = 0.9\linewidth]{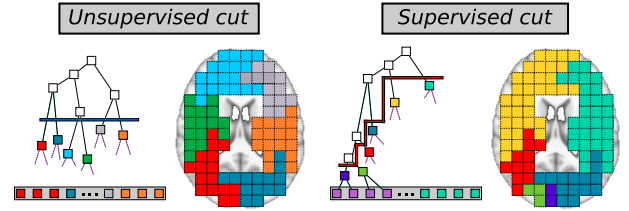}
\caption{\emph{Top-Down step (Pruning of the tree)} - step \ref{Step:pruning}.
In the \emph{unsupervised cut} approach, (left) Ward's tree is divided
into $6$ parcels through a horizontal cut (blue). In the \emph{supervised
cut} approach (right), by choosing the best cut (red) of the tree given a
score   function $\zeta_e$, we focus on some specific regions of the tree
that are more informative.}
\label{Fig:ward_cut}
\end{figure}

We now detail how the tree $\mathcal{T}$ can be pruned to create a reduced set
of \emph{parcellations}.
Because the hierarchical subdivision of the brain volume (by
successive inclusions) is naturally identified as a tree $\mathcal{T}$, choosing
a
parcellation adapted to the prediction problem means optimizing a cut of the
tree. Each sub-tree created by the cut represents a region whose
average signal is used for prediction.
As no optimal solution is currently available to solve this problem,
we consider two approaches to perform such a cut (see
Fig.~\ref{Fig:ward_cut}). In order to have $\Delta$ parcels, these two methods
start from the root of the tree
$\mathcal{T}$ (one unique parcel for the whole brain), and iteratively refine
the parcellation:

\begin{itemize}
\item The first solution consists in using the \emph{inertia criterion} from
Ward algorithm: the cut consists in a subdivision of the Ward's tree into its
$\Delta$ main branches. As this does not take into account the target
information $\bold{y}$, we call it \emph{unsupervised cut (UC)}.

\item The second solution consists in initializing the cut at the highest
level of the hierarchy and then successively finding the new sub-tree
cut that maximizes a prediction score $\zeta$ (\emph{e.g.} explained variance,
see Eq.(\ref{Eq:ev}) below),
while using a prediction function $\mathcal{F}$  (\emph{e.g.} Support Vector
Machine \cite{cortes1995}) instantiated with the parcels-based signal averages
at the current step. As in a greedy
approach,
successive cuts iteratively create a finer parcellation of the search
volume, yielding the set of parcellations
$\mathcal{P}_{1},..,\mathcal{P}_{\Delta}$. More specifically, one parcel
is split at each step, where
the choice of the split is driven by the prediction problem. After $\delta$ such
steps of exploration, the brain
is divided into
$\delta+1$ parcels. This procedure, called \emph{supervised cut (SC)}, is
detailed in algorithm \ref{Tab:PseudoCode_sc}.

\end{itemize}

\subsubsection{Model Selection step: optimal sub-tree $\widehat{\mathcal{T}}$}
\label{Step:sel}

In both cases, a set of nested parcellations  is produced, and the
optimal model among the available cuts still has to be chosen. 
We select the sub-tree $\widehat{\mathcal{T}}$ that yields the optimal
prediction score $\widehat{\zeta}$.
The corresponding optimal parcellation is then used to create parcels on
both training and test sets. A prediction function is thus
trained and tested on these two set of parcels to compute the prediction
accuracy of the framework.

\subsection{Algorithmic considerations}

The \emph{pruning of the tree} and the \emph{model selection} step are
included in an internal cross-validation procedure within the training
set. However, this internal cross-validation scheme rises different
issues. First, it is very time consuming to include the two steps
within a complete internal cross-validation.  A second, and more
crucial issue, is that performing an internal cross-validation over
the two steps yields many different sub-trees (one by fold).  However,
it is not easy to combine these different sub-trees in order to obtain
an average sub-tree that can be used for prediction on the test set
\cite{oliver1995}. Moreover, the different optimal sub-trees are not
constructed using all the training set, and thus depend on the
internal cross-validation scheme.  Consequently, we choose an
empirical, and potentially biased, heuristic that consists of using
sequentially two separate cross-validation schemes $C_e$ and $C_s$ for
the \emph{pruning of the tree} and the \emph{model selection} step.

\subsection{Computational considerations}

Our algorithm can be used to search informative regions in very
high-dimensional data, where other algorithms do not scale well.
Indeed, the highest number of features considered by our approach is $\Delta$,
and we can use any given prediction function $\mathcal{F}$, even if this
function is not well-suited for high dimensional data.
The computational complexity of the proposed \emph{supervised clustering}
algorithm depends thus on the complexity of the prediction
function $\mathcal{F}$, and on the two cross-validation schemes $C_e$  and
$C_s$.
At the current iteration $\delta \in [1,\Delta]$, $\delta+1$ possible
features are considered in the regression model, and the regression function
is fit $n(\delta+1)$ times (in the case of a leave-one-out cross-validation
with $n$ samples). Assuming the cost of fitting the prediction function
$\mathcal{F}$ is $\mathcal{O}(\delta^\alpha)$ at step $\delta$, the overall
cost complexity of the procedure is
$\mathcal{O}(n\Delta^{(2+\alpha)})$. In general $\Delta \ll p$, and the cost
remains affordable as long as $\Delta<10^3$, which was the case in all our
experiments. Higher values for $\Delta$ might also be used, but the
complexity of $\mathcal{F}$ has to be lower.

The benefits of parcellation come at a cost regarding CPU time. On a subject of
the dataset on the prediction of size (with a non optimized Python
implementation though), with $\sim 7. 10^{4}$ voxels, the
construction of the tree raising CPU time to $207$ seconds and the parcels
definition raising CPU time (\emph{Intel(R) Xeon(R), 2.83GHz}) to $215$
seconds.
Nevertheless, all this remains perfectly affordable for standard
neuroimaging data analyzes.

\begin{algorithm}[h!tb]
\caption{Pseudo-code for \emph{supervised cut}}
\textbf{Set} a number of exploration steps $\Delta$, a score function $\zeta$, a
prediction function $\mathcal{F}$, and two cross-validation schemes $C_e$
and $C_s$.\\
\textbf{Let} $\mathcal{P}_{\delta}$ be the parcellation defined at the current
iteration $\delta$ and $\bold{X_p}_\delta$ the corresponding
\emph{parcel-based}
averages.\\
\textbf{Construct} $\mathcal{T}$ using Ward algorithm.\\
\textbf{Start} from the root of the tree $\mathcal{T}$, \emph{i.e.}
$\mathcal{P}_{0} = \{P_0\}$ has only one parcel $P_0$ that contains all the
voxels.\\

\emph{Pruning of the tree $\mathcal{T}$}\\
\For{$\delta \leftarrow 1$ \KwTo $\Delta$}{
        \ForEach{$P_i \in \mathcal{P}_{\delta-1}$}{
            - Split $P_i \rightarrow \{P_i{^1}, P_i{^2}\}$ according to
                        $\mathcal{T}$.\\
            - Set $\mathcal{P}_{\delta,i} = \{\mathcal{P}_{\delta-1} \backslash
                    P_i\} \cup  \{P_i{^1}, P_i{^2}\}$.\\
            - Compute the corresponding \emph{parcel-based} signal averages
                            $\bold{X_p}_{\delta,i}$.\\
            - Compute the cross-validated score $\zeta_{e,i}(\mathcal{F})$ with
                the cross-validation scheme $C_e$.
            }
        - Perform the split $i^\star$ that yields the highest score
            $\zeta_{e,i^\star}(\mathcal{F})$.\\
        - Keep the corresponding parcellation $\mathcal{P}_{\delta}$ and
sub-tree $\mathcal{T}_{\delta}$.}

\emph{Selection of the optimal sub-tree $\widehat{\mathcal{T}}$}\\
\For{$\delta \leftarrow 1$ \KwTo $\Delta$}{
        - Compute the cross-validated score $\zeta_{s,\delta}(\mathcal{F})$
with the cross-validation scheme $C_s$, using the parcellation
$\mathcal{P}_{\delta}$.
            }

\textbf{Return} the sub-tree $\widehat{\mathcal{T}}_{\delta^\star}$ and
corresponding parcellation $\widehat{\mathcal{P}}_{\delta^\star}$, that yields
the highest score $\zeta_{s,\delta^\star}(\mathcal{F})$.

\label{Tab:PseudoCode_sc}
\end{algorithm}

\subsection{Performance evaluation}

Our method is evaluated with a cross-validation procedure that splits
the available data into training and validation sets.
In the following, $(\bold{X}^{l},\bold{y}^{l})$
are a learning set, $(\bold{X}^{t},\bold{y}^{t})$ a test set and
$\bold{\hat{y}}^{t}=f(\bold{X}^t\bold{\hat{w}})$ refers to the predicted target,
where $\bold{\hat{w}}$
is estimated from the training set.
For regression analysis, the performance of the different models is
evaluated using $\zeta$, the ratio of explained variance:
\begin{equation}
\zeta(\bold{y}^{t},\bold{\hat{y}}^{t}) = \frac{\mbox{var}(\bold{y}^{t}) -
\mbox{var}\left(\bold{y}^{t} -
\bold{\hat{y}}^{t}\right)}{\mbox{var}(\bold{y}^{t})}
\label{Eq:ev}
\end{equation}
This is the amount of variability in the response that can be
explained by the model. A perfect prediction yields $\zeta = 1$, a constant
prediction yields $\zeta=0$).
For classification analysis, the performance of the different models is
evaluated using a standard classification score denoted $\kappa$ , defined
as:
\begin{equation}
\kappa(\bold{y}^{t},\bold{\hat{y}}^{t}) =
\frac{\sum_{i=1}^{n^t}\delta(y^{t}_i,\hat{y}_i^{t})}{n^t}
\end{equation}
where $n^t$ is the number of samples in the test set, and $\delta$ is
Kronecker's delta.

\subsection{Competing methods}
In our experiments, the \emph{supervised clustering} is compared to different
state of the art regularization methods. For regression experiments:
\begin{itemize}
\item \emph{Elastic net} regression \cite{zou2005}, requires
setting two parameters $\lambda_1$ (amount of $\ell_1$ norm regularization) 
and $\lambda_2$ (amount of $\ell_2$ norm regularization). In our analyzes, an
internal cross-validation procedure on the training set is
used to optimize $\lambda_1 \in \{0.2
\tilde{\lambda},0.1 \tilde{\lambda},0.05
\tilde{\lambda}, 0.01
  \tilde{\lambda}\}$, where $\tilde{\lambda} = \|\bold{X}^T
\bold{y}\|_\infty$, and
  $\lambda_2  \in \{0.1,0.5,1.,10.,100.\}$.
\item \emph{Support Vector Regression} (\emph{SVR}) with a linear
  kernel \cite{cortes1995}, which is the reference method in
  neuroimaging. The regularization  parameter $C$ is optimized by
cross-validation in the range of $10^{-3}$ to
  $10$ in multiplicative steps of $10$.
\end{itemize}

For classification settings:
\begin{itemize}
\item \emph{Sparse multinomial logistic regression} (\emph{SMLR})
  classification \cite{krishnapuram2007}, that requires an optimization similar
  to \emph{Elastic Net} (two parameters $\lambda_1$ and $\lambda_2$).
\item \emph{Support Vector Classification} (\emph{SVC}), which is optimized 
similarly as \emph{SVR}.

\end{itemize}

All these methods are used after an \emph{Anova}-based
feature selection as this maximizes their performance. Indeed, irrelevant
features and redundant information can decrease the accuracy of a predictor
\cite{hughes1968}.
This selection is performed on the training set, and the optimal number of
voxels is selected in the range $\{50,100,250,500\}$ within a nested
cross-validation. 
We also check that increasing the range
of voxels (\emph{i.e.} adding 2000 in the range of number of selected voxels)
does not increase the prediction accuracy on our real datasets.
The implementation of \emph{Elastic net} is based on \emph{coordinate
descent} \cite{friedman2009}, while \emph{SVR} and \emph{SVC} are based on 
LibSVM \cite{libsvm}. Methods are used from \emph{Python} via the
\emph{Scikit-learn} open source package~\cite{scikit}.
Prediction accuracies of the different methods are compared using a paired
t-test.


\section{Simulated data}
\label{sec:simu}

\subsection{Simulated one-dimensional data}

We illustrate the \emph{supervised clustering} on a simple simulated data
set, where the informative features have a block structure:
%
\begin{equation}
\bold{X} \sim \mathcal{N}(0,1)
\;\;\mbox{and}\;\;
\bold{y} =  \bold{X}\bold{w} + \bold{\epsilon}
\end{equation}
with $\bold{\epsilon} \sim \mathcal{N}(0,1)$ and $\bold{w}$ is defined as $w_i
\sim \mathcal{U}_{0.75}^{1.25}$ for $20 \leq i \leq 30$, $w_i \sim
\mathcal{U}_{-1.25}^{-0.75}$ for $50 \leq i \leq 60$, and $w_i = 0$
elsewhere, where $ \mathcal{U}_{a}^{b}$ is the uniform distribution between $a$
and $b$.
We have $p=200$ features and $n=150$ images. The \emph{supervised cut} is 
used with $\Delta=50$, \emph{Bayesian Ridge Regression (BRR)} as prediction
function $\mathcal{F}$, and procedures $C_e$ and $C_s$ are set to 4-fold
cross-validation.

\subsection{Simulated neuroimaging data}

The simulated data set $\bold{X}$ consists in $n = 100$ images (size
$12\times12\times12$ voxels) with a set of four cubic Regions of
Interest (ROIs) (size $2\times2\times2$). We call $\mathcal{R}$ the
support of the ROIs (\emph{i.e.}  the $32$ resulting voxels of
interest). Each of the four ROIs has a fixed weight in
$\{-0.5,0.5,-0.5,0.5\}$. We call $w_{i,j,k}$ the weight of the
$(i,j,k)$ voxel.
To simulate the spatial variability between images (inter-subject variability,
movement artifacts in intra-subject variability), we define a new support of
the ROIs, called $\tilde{\mathcal{R}}$ such as, for each image, half (randomly
chosen) of the weights $\bold{w}$ are set to zero. Thus, we
have $\tilde{\mathcal{R}} \subset \mathcal{R}$.
%
%
We simulate the signal in the $(i,j,k)$ voxel of the $l^{th}$ image as:
\begin{equation}
\bold{X}_{i,j,k,l} \sim \mathcal{N}(0,1)
\label{eq:sc_sim2}
\end{equation}
The resulting images are smoothed with a Gaussian kernel with a standard
deviation of $2$ voxels, to mimic the correlation structure observed in real
fMRI data.
The target $\bold{y}$ for the $l^{th}$ image is simulated as:
\begin{equation}
\bold{y}_{l} = \sum_{(i,j,k) \in \tilde{\mathcal{R}}} w_{i,j,k}
\bold{X}_{i,j,k,l} +
\epsilon_l
\label{eq:sc_sim1}
\end{equation}
and  $\epsilon_l \sim \mathcal{N}(0,\gamma)$ is a Gaussian noise with
standard deviation $\gamma > 0.$
We choose $\gamma$ in order to have a signal-to-noise (SNR) ratio of $5$
dB. The SNR is defined here as $20$ times the log of the ratio between the
norm of the signal and the norm of the added noise.
We create a training set of $100$ images, and then we validate on
$100$ other images simulated according to
Eq.~\ref{eq:sc_sim2}-\ref{eq:sc_sim1}.
We compare the \emph{supervised clustering} approach
with the \emph{unsupervised clustering}  and the two reference
algorithms, \emph{Elastic net} and \emph{SVR}. The two reference methods are
optimized by 4-fold cross-validation within the training set in the range
described below.
We also compare the methods to a \emph{searchlight} approach
\cite{kriegeskorte2006} (radius of $2$ and $3$ voxels, combined with
a \emph{SVR} approach ($C=1$)), which has emerged as a reference approach for
decoding local fine-grained information within the brain.

Both \emph{supervised cut} and \emph{unsupervised cut} algorithms  are used
with $\Delta = 50$, \emph{Bayesian Ridge Regression (BRR)} as prediction
function $\mathcal{F}$, and optimized with an internal 4-fold cross-validation.

\subsection{Results on one-dimensional simulated data}

The results of the \emph{supervised clustering} algorithm are given
in Fig.~\ref{Fig:simu_1d}. On the top, we give the tree $\mathcal{T}$, where
the parcels found by the \emph{supervised clustering} are represented by red
squares, and the bottom row are the input features. The features of interest
are represented by green dots. We note that the algorithm focuses the
parcellation on two sub-regions, while leaving other parts of the tree
unsegmented.
The weights found by the prediction function based on the optimal parcellation
(bottom) clearly outlines the two simulated informative regions.  The predicted
weights are normalized by the number of voxels in
each parcel.

\subsection{Results on simulated neuroimaging data}

\begin{figure}[h!tb]
\center \includegraphics[width = 1.\linewidth]{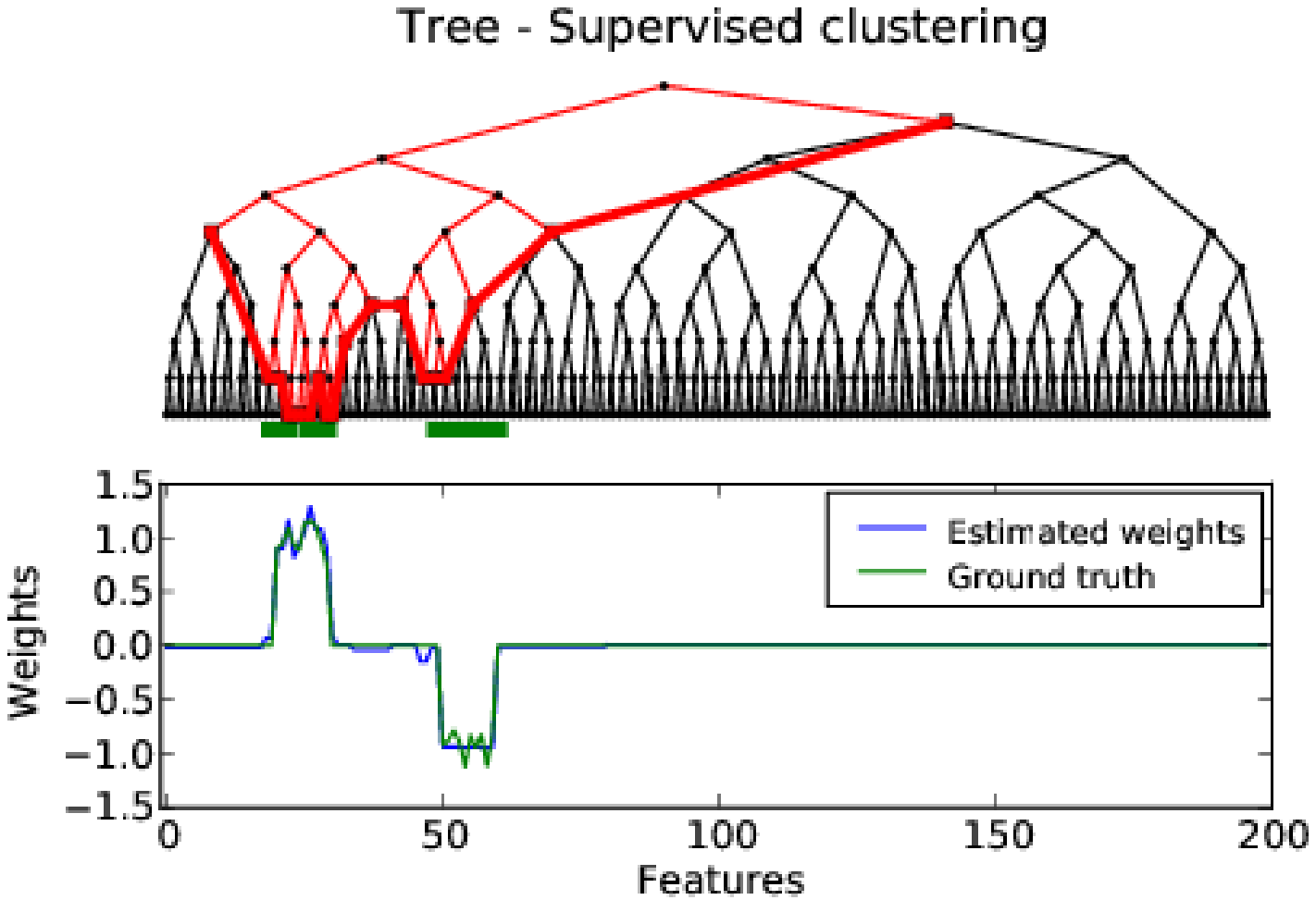}
\caption{Illustration of the \emph{supervised clustering} algorithm on a simple
simulated data set. The cut of the tree (top, red line) focuses on the regions
of interest (top, green dots), which allows the prediction function to
correctly weight the informative features (bottom).}
\label{Fig:simu_1d}
%
%



\begin{center}
\begin{minipage}{0.24\linewidth}
\center \includegraphics[width = 1.\linewidth]{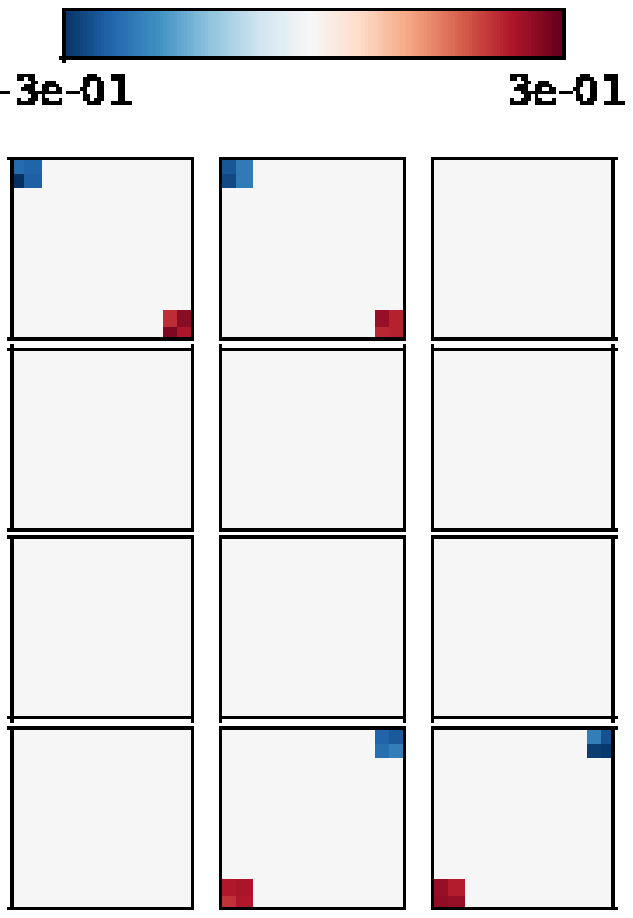}
\center \scriptsize \vspace{-3ex} \emph{(a) True\\weights}
\end{minipage}
\hfill
\begin{minipage}{0.24\linewidth}
\center \includegraphics[width = 1.\linewidth]{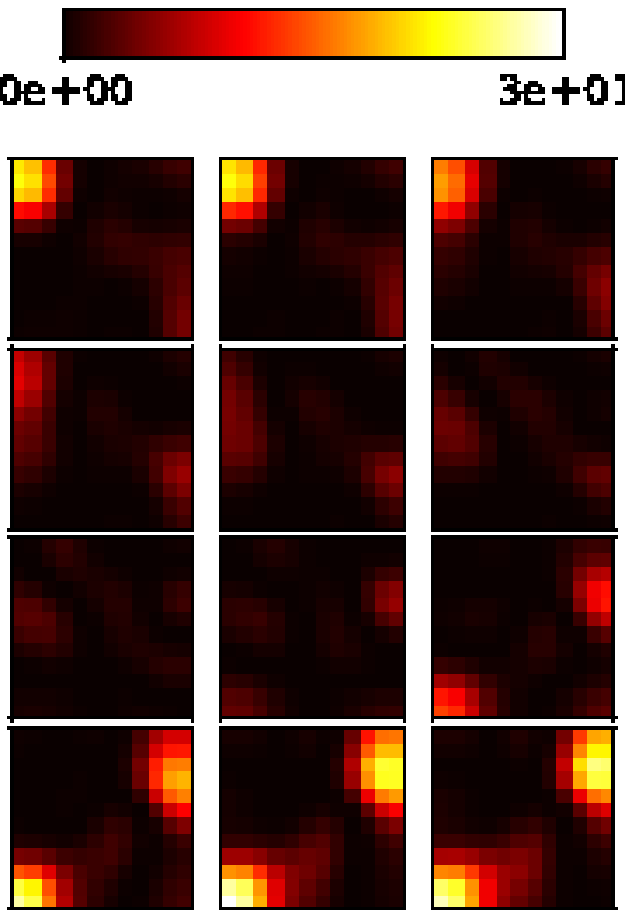}
\center \scriptsize \vspace{-3ex} \emph{(b) Anova\\F-scores}
\end{minipage}
\hfill
\begin{minipage}{0.24\linewidth}
\center \includegraphics[width = 1.\linewidth]{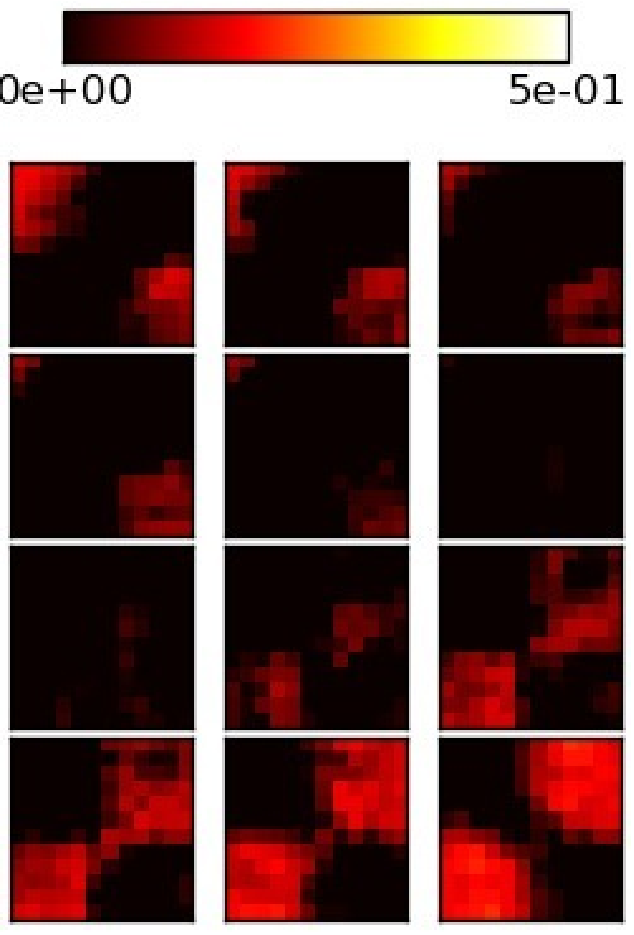}
\center \scriptsize \vspace{-3ex} \emph{(c) Searchlight\\SVR ($r = 2$)}
\end{minipage}
\hfill
\begin{minipage}{0.24\linewidth}
\center \includegraphics[width = 1.\linewidth]{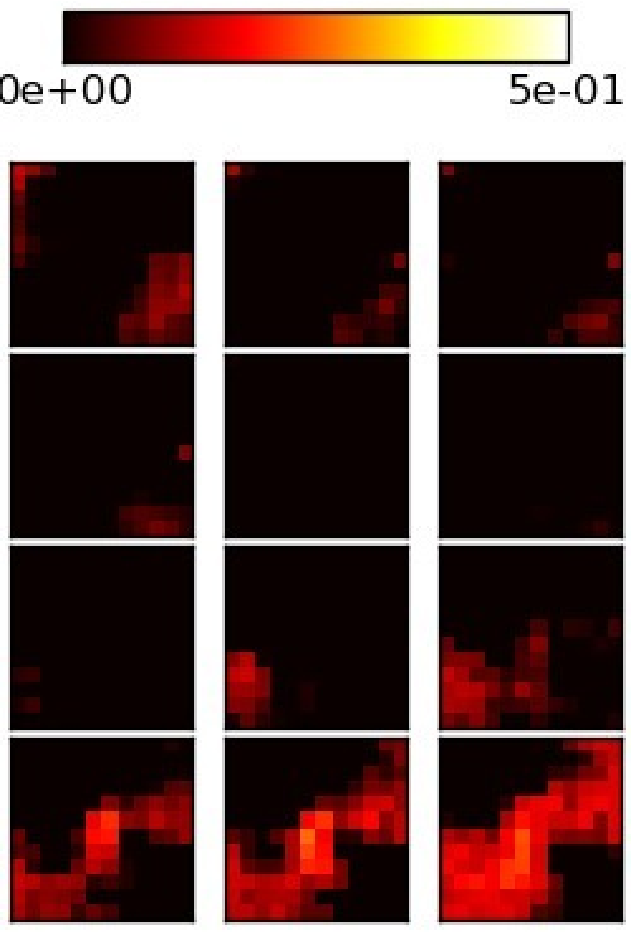}
\center \scriptsize \vspace{-3ex} \emph{(d) Searchlight\\SVR ($r = 3$)}
\end{minipage}

\vspace{1ex}


\begin{minipage}{0.24\linewidth}
\center \includegraphics[width = 1.\linewidth]{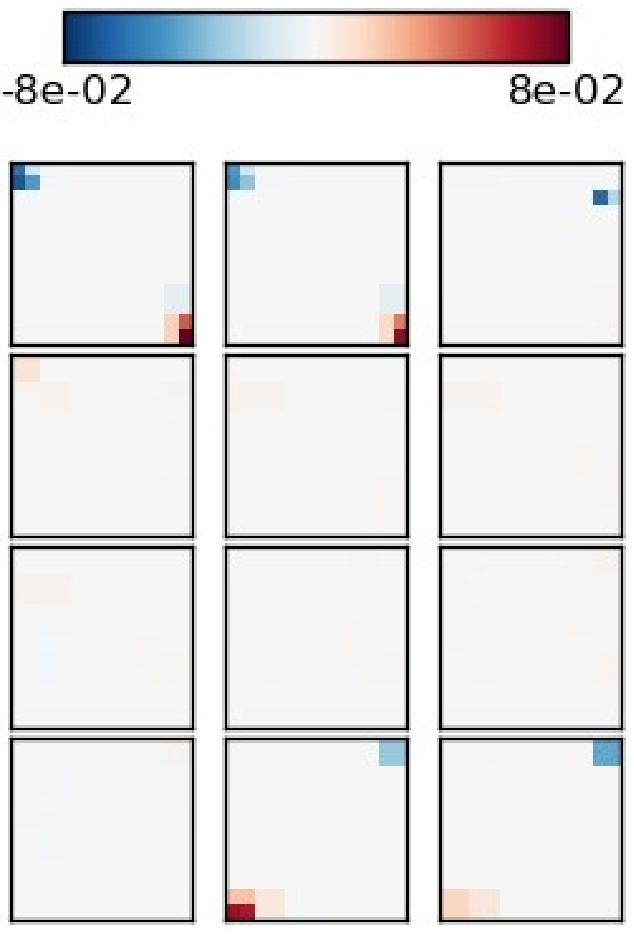}
\center \scriptsize \vspace{-3ex} \emph{(e) Supervised\\clustering
}
\end{minipage}
\hfill
\begin{minipage}{0.24\linewidth}
\center \includegraphics[width = 1.\linewidth]{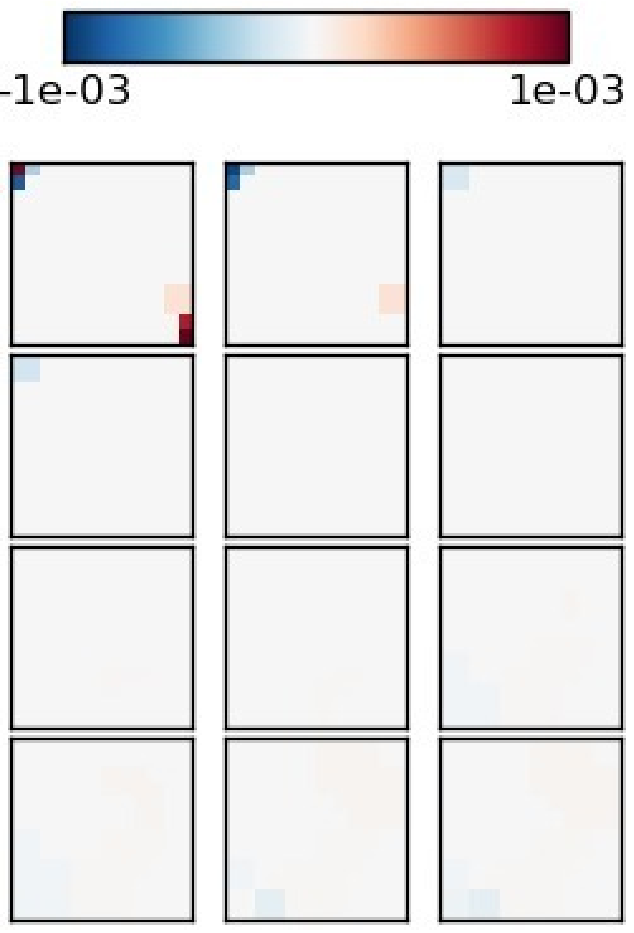}
\center \scriptsize \vspace{-3ex} \emph{(f) Unsupervised\\clustering
}
\end{minipage}
\hfill
\begin{minipage}{0.24\linewidth}
\center \includegraphics[width = 1.\linewidth]{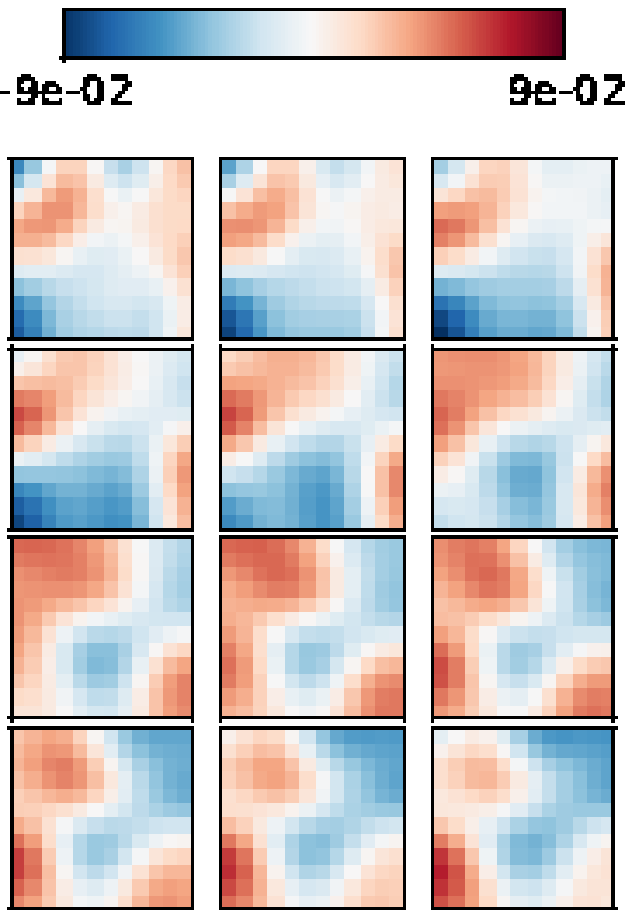}
\center \scriptsize \vspace{-3ex} \emph{(g) SVR\\Cross-validated
}
\end{minipage}
\hfill
\begin{minipage}{0.24\linewidth}
\center \includegraphics[width = 1.\linewidth]{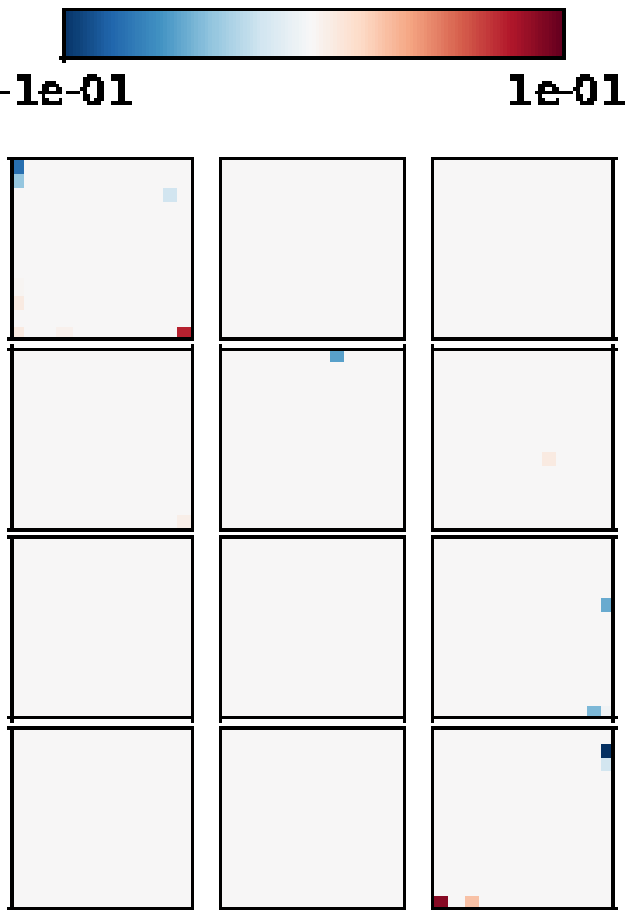}
\center \scriptsize \vspace{-3ex} \emph{(h) Elastic net\\Cross-validated
}
\end{minipage}

\end{center}


\begin{center}
\caption{
Comparisons of the weights given by the different procedures (b-h)
with the true weights (a). Only the \emph{supervised cut} algorithm (e)
retrieves the regions of interest.
For the searchlight approach (c, f), the images show the explained variance
obtained using the voxels within a sphere centered on each voxel.}
\label{fig:ResSimuParam}
\end{center}
\vspace{-6ex}
\end{figure}

We compare different methods on the simulated data, see
Fig.~\ref{fig:ResSimuParam}.
The predicted weights of the two parcel-based approaches are normalized by the
number of voxels in each parcel.
Only the \emph{supervised clustering} (e)
extracts the simulated discriminative regions. The \emph{unsupervised
clustering} (f) does not retrieve the whole support of the weights, as the
created \emph{parcels} are constructed based only on the signal and spatial
information, and thus do not consider the target to be predicted.
\emph{Elastic net} (h) only retrieves part of the support of the weights, and
yields an overly sparse solution which is not easy to interpret. \emph{SVR}
(g) approach yields weights in the primal space that depend on the
smoothness of the images.
The searchlight approach (c,d), which is a commonly used brain mapping
techniques,
shows here its limits: it does not cope with the long range multivariate
structure of the weights, and yields very blurred informative maps, because
this method naturally degrades data resolution.

\section{Experiments and results on real data}
\label{sec:results}

\subsection{Details on real data}

We apply the different methods to analyze ten subjects from an fMRI dataset
related to the study
of the visual representation of objects in the brain (see \cite{eger2007} for
details).
During the experiment, ten healthy volunteers viewed objects of two categories
(each one of the two categories is used in half of the subjects)
with four different exemplars in each category. Each exemplar was presented at 
three different sizes (yielding
$12$ different experimental conditions per subject). Each stimulus was
presented four times in each of the six sessions. We averaged data from the four
repetitions, resulting in a total of $n=72$ images by subject (one image of
each stimulus by session).
Functional images were acquired on a 3-T MR system with eight-channel
head coil (Siemens Trio, Erlangen, Germany) as T2*-weighted
echo-planar image (EPI) volumes. Twenty transverse slices were
obtained with a repetition time of 2s (echo time, 30ms; flip angle,
$70^{\circ}$; $2\times2\times2$-mm voxels; $0.5$-mm gap).
Realignment, normalization to MNI space, and General Linear Model (GLM)
fit were performed with the SPM5 software\footnote{http://
www.fil.ion.ucl.ac.uk/spm/software/spm5}.
In the GLM, the time course of each of the $12$ stimuli convolved with a
standard
hemodynamic response function was modeled separately, while
accounting for serial auto-correlation with an AR(1) model and
removing low-frequency drift terms with a high-pass filter with a
cut-off of 128 s. In the present work we used
the resulting session-wise parameter estimate images.
All the analysis are performed on the whole brain volume.

\textbf{Regression experiments:} The four different exemplars in
each of the two categories were pooled, leading to images labeled according to
the 3 possible sizes of the object. By doing so, we are
interested in finding discriminative information to predict the size 
of the presented object. This reduces to a regression problem, in which our
goal
is to predict a
simple scalar factor (size or scale of the presented object).

We perform an inter-subject regression analysis on the sizes.  This analysis
relies on subject-specific fixed-effects activations, \emph{i.e.} for
each condition, the six activation maps corresponding to the six
sessions are averaged together.  This yields a total of twelve images
per subject, one for each experimental condition. The dimensions
of the real data set are $p \sim 7\times 10^4$ and $n=120$ (divided into three
different sizes).
We evaluate the performance of the method by
cross-validation (leave-one-subject-out).
The parameters of the reference methods are optimized with a
nested leave-one-subject-out cross-validation within the training set, in the
ranges given before.
The \emph{supervised clustering} and \emph{unsupervised clustering} are used
with \emph{Bayesian Ridge Regression (BRR)} (as described in section
3.3 in \cite{bishop2007}) as prediction function $\mathcal{F}$. Internally, a
\emph{leave-one-subject-out}
cross-validation is used and 
we set the maximal number of parcels to $\Delta=75$. The optimal number of
parcels is thus selected between $1$ and $75$ by a nested cross-validation
loop.

A major asset of \emph{BRR} is
that it adapts the
regularization to the data at hand, and thus can cope with the different
dimensions of the problem: in the first steps of the \emph{supervised
clustering} algorithm, we have more samples than features, and for the last
steps, we have more features than samples.
The two hyperparameters that governed the gamma distribution of the
regularization term of BRR are both set to $10^{-6}$ (the prior is weakly
informative). We do not optimize these hyperparameters, due to computational
considerations, but we check that with more informative priors we obtain
similar results in the regression experiment ($0.81$ and $0.79$ with
respectively $\lambda_1 = \lambda_2 = 0.01$ and $\lambda_1 = \lambda_2 = 1.$).

\textbf{Classification experiments:} We evaluate the performance on a second
type of discrimination which is object classification. In that case, we averaged
the images for the three
sizes and we are interested in discriminating between individual object
shapes. For each of the two categories, this can be
handled
as a classification problem, where we aim at predicting the shape of
an object corresponding to a new fMRI scan. We perform two analyses
corresponding to the two categories used, each one including five subjects.

In this experiment, the \emph{supervised clustering} and \emph{unsupervised
clustering}  are used with \emph{SVC} ($C=0.01$) as prediction function
$\mathcal{F}$.
Such value of $C$ yields a good regularization of the weights in the
proposed approach, and the results are not too sensitive to this parameter
($67.5 \%$ for $C=10$)

\subsection{Results for the prediction of size}

\begin{figure}[h!tb]
 \center \includegraphics[width=1.\linewidth]{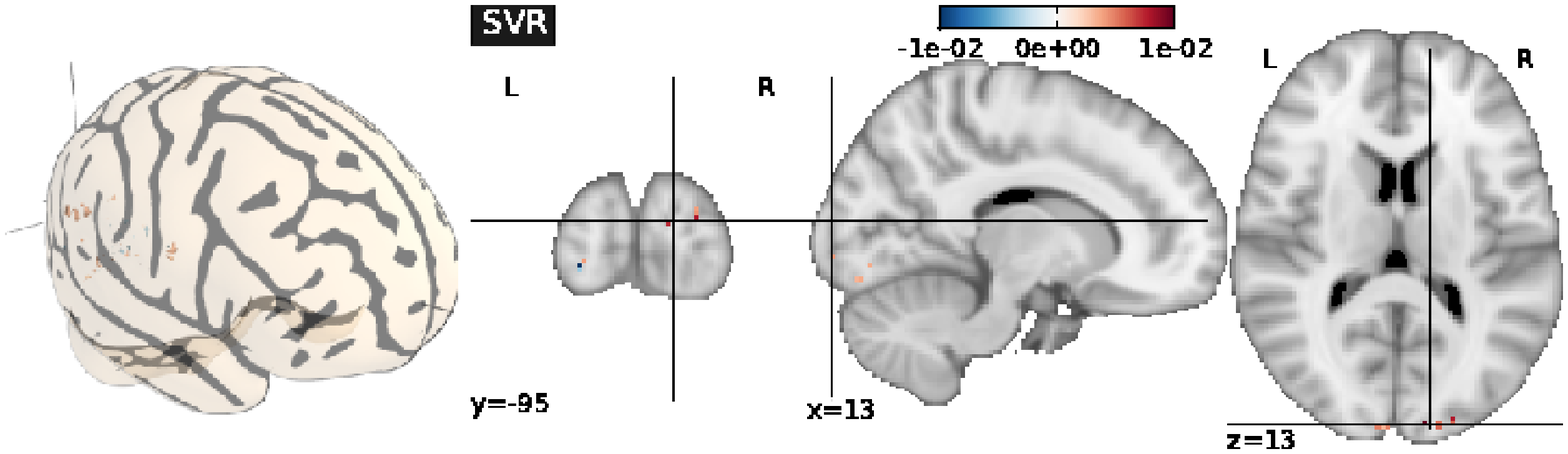}
\vspace{-2ex}
 \center \includegraphics[width=1.\linewidth]{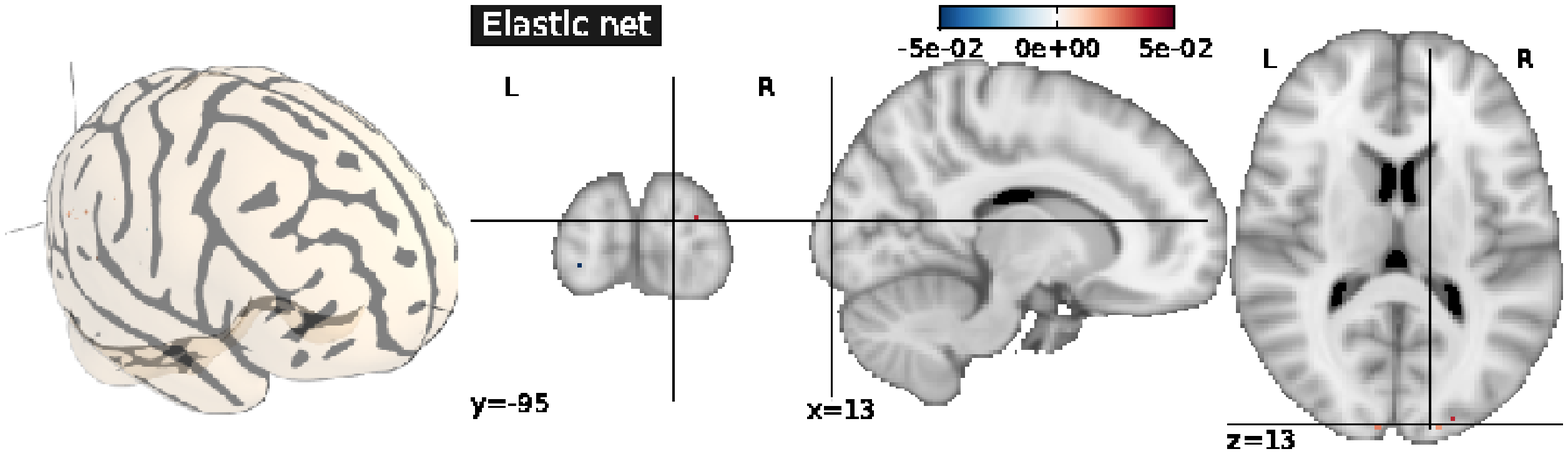}
\vspace{-2ex}
\center \includegraphics[width=1.\linewidth]{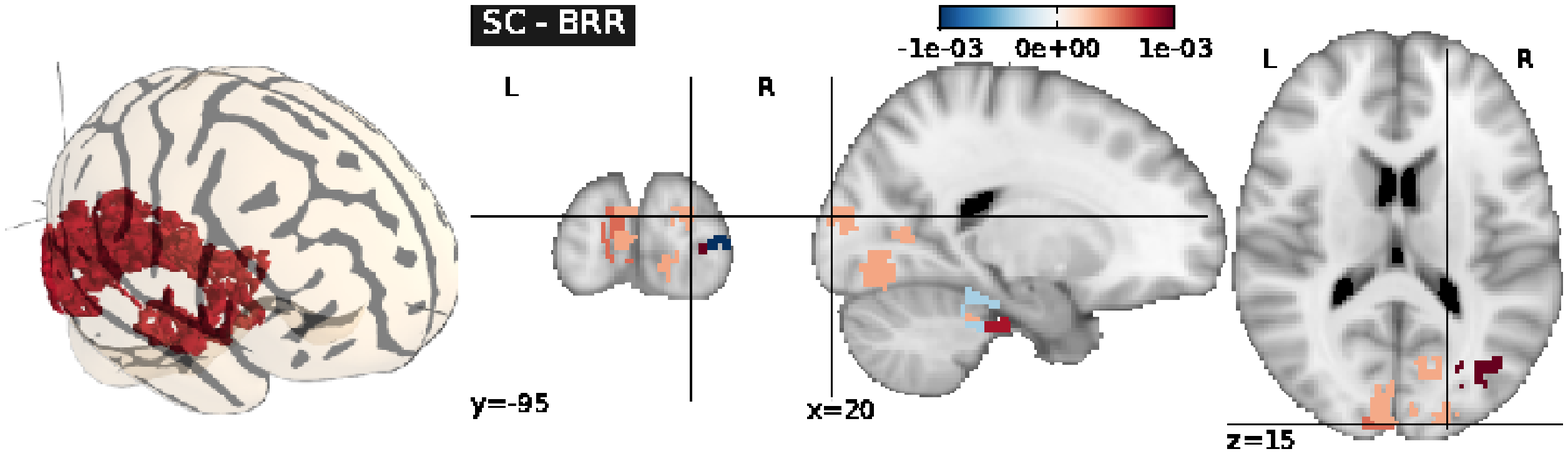}
\vspace{-2ex}
\center \includegraphics[width=1.\linewidth]{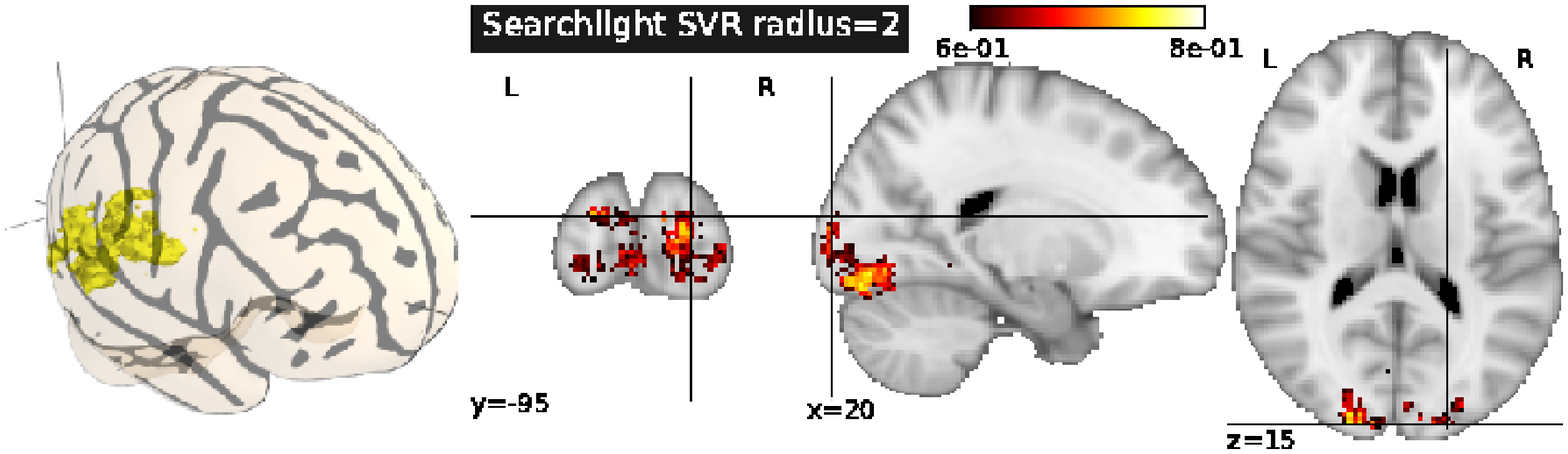}
\vspace{-2ex}
\caption{\emph{Results for prediction of size}.
Maps of weights found by \emph{supervised cut}, the two reference
voxel-based methods and the searchlight. The proposed algorithm creates very
interpretable clusters, compared to the reference methods, which is related
to the fact that they do not consider the spatial structure of the image.
Moreover, the \emph{supervised clustering} yields similar maps as
\emph{searchlight}, but also retrieves some additional clusters.}
\label{Fig:real_w_sizes_inter}
\end{figure}

The results of the inter-subjects analysis are given in
Tab.\ref{Tab:res_sizes_inter_reg}.
Both parcel-based methods perform better than voxel-based reference
methods.
Parcels can be seen as an accurate method for compressing information without
loss of prediction performance.
%
Fig. \ref{Fig:real_w_sizes_inter} gives the weights found for the
\emph{supervised cut}, the two reference methods and the searchlight ($SVR$ with
$C=1$ and a radius of $2$ voxels), using the whole data set.
As one can see, the
proposed algorithm yields clustered loadings map, compared to the
maps yielded by the voxel-based methods, which are very sparse and difficult to
represent.
Compared to the searchlight, the \emph{supervised clustering} creates more
clusters that are also easier to interpret as they are  well separated.
Moreover, the proposed approach yields a prediction accuracy for the whole
brain analysis, a contrario to the searchlight that only gives a local measure
of information.

The majority of informative
parcel are located in the posterior part of the occipital cortex,  most likely
corresponding to primary visual cortex, with few additional slightly more
anterior parcels in posterior lateral occipital cortex. This is consistent with
the previous findings \cite{eger2007} where a gradient of sensitivity to size
was
observed across object selective lateral occipital \emph{ROIs}, while the
most accurate discrimination of sizes is obtained in primary visual cortex.

\begin{table}[h!tb]
\begin{center}
\footnotesize{
\begin{tabular}{|l|c|c|c|c|c|}
\hline
 Methods & mean $\zeta$ & std $\zeta$ & max $\zeta$ & min $\zeta$ & p-val to
UC\\
\hline 
\hline 
SVR & $0.77$ & $0.11$ & $0.97$ & $0.58$ & $0.0817$\\ 
\hline 
Elastic net & $0.78$ & $0.1$ & $0.97$ & $0.65$ & $0.0992$ \\
\hline 
\rowcolor{lightgray}UC - BRR & $0.83$ & $0.08$ & $0.97$ & $0.73$ & -  \\ 
\hline 
SC - BRR & $0.82$ & $0.08$ & $0.93$ & $0.7$ & $0.8184$ \\ 
\hline 
\end{tabular}
}
\end{center}
\caption{Explained variance $\zeta$
for the different methods in the \emph{Size prediction analysis}. The
p-values are computed using a paired t-test. The
\emph{unsupervised cut (UC)} algorithm yields the
best prediction accuracy (leave-one-subject-out cross-validation). The
\emph{supervised cut (SC)} yields similar
results as \emph{UC} (the difference is not significant).
The two voxel-based approaches yield lower prediction accuracy than parcel-based
approaches.}
\label{Tab:res_sizes_inter_reg}
%
%


\begin{center}
\footnotesize{
\begin{tabular}{|l|c|c|c|c|c|}
\hline
 Methods & mean $\kappa$ & std $\kappa$ & max $\kappa$ & min $\kappa$ & p-val
to SC  \\
\hline
\hline 
SVC & $48.33$ & $15.72$ & $75.0$ & $25.0$ & $0.0063$ **\\ 
\hline 
SMLR & $42.5$ & $9.46$ & $58.33$ & $33.33$ & $0.0008$ **\\ 
\hline 
UC - SVC & $65.0$ & $8.98$ & $75.0$ & $50.0$ & $0.1405$ \\ 
\hline 
\rowcolor{lightgray}SC - SVC & $70.0$ & $10.67$ & $83.33$ & $50.0$ & - \\ 
\hline 
\end{tabular}
}

\vspace{2ex}

\footnotesize{
\begin{tabular}{|l|c|c|c|}
\hline
 Methods & mean (std) $\kappa$& mean (std) $\kappa$& Mean nb.
feat.\\
  & cat. 1& cat.2 &  (voxels/parcels)\\
\hline
\hline 
SVC & $56.6 (17.8)$ & $40.0 (6.2)$ & $415$\\ 
\hline 
SMLR & $43.3 (9.7)$ & $41.6 (9.1)$ & $150$\\ 
\hline 
UC - SVC & $63.3 (8.5)$ & $68.3 (9.7)$ & $21$ \\ 
\hline 
\rowcolor{lightgray}SC - SVC & $65 (12.2)$ & $75 (5.2)$ & $17$\\ 
\hline 
\end{tabular}
}
\end{center}
\caption{Top -- Classification performance $\kappa$ for the different
methods in the
\emph{Object prediction analysis}. The
p-values are computed using a paired t-test. The
\emph{supervised cut (SC)} algorithm yields the
best prediction accuracy (leave-one-subject-out cross-validation).
Both parcels-based approaches are
significantly more accurate and more stable than voxel-based approaches.
Bottom -- Details of the results for the two categories and mean number of
features (voxels or parcels) for the different methods. We can notice that
parcels yield a good compression of information has with more than ten times
less features, parcel-based approaches yield higher prediction accuracy.}
\label{Tab:res_objects_inter_classif}

\end{table}

\subsection{Results for the prediction of shape}

The results of the inter-subjects analysis are given
in Tab.\ref{Tab:res_objects_inter_classif}.
The \emph{supervised cut} method outperforms the other approaches. In
particular, the classification score is $21\%$ higher than with
voxel-based \emph{SVC} and $27\%$ higher than with voxel-based \emph{SMLR}.
Both parcel-based approaches are
significantly more accurate and more stable than the voxel-based approaches.
The number of features used show the good compression of information
performed by the parcels. With ten times less features than voxel-based
approaches, the prediction accuracies of parcel-based approaches are higher.
The lower performances of \emph{SVC} and \emph{SMLR} can be explained by the
fact that voxel-based approaches can not deal with inter-subject variability,
especially in such cases where information can be encoded in pattern of voxels
that can vary spatially across subjects.

\section{Discussion}
\label{sec:discussion}

In this paper, we have presented a new method for enhancing the prediction of
experimental variables from \emph{fMRI} brain images.
The proposed approach constructs \emph{parcels} (groups of connected
voxels) by \emph{feature agglomeration} within the whole brain, and allows
to
take into account both the spatial structure and the multivariate
information within the whole brain.

\begin{figure}[h!tb]
\begin{center}
\includegraphics[width=0.7\linewidth]
{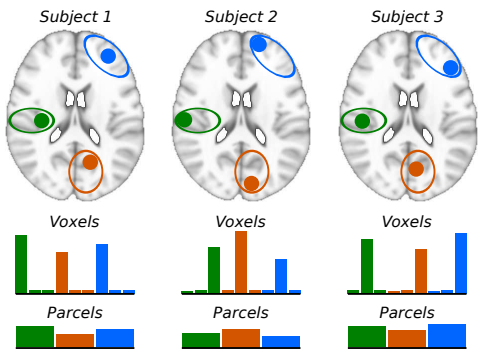}
\caption{Illustration of \emph{feature agglomeration} to cope with
inter-subject
variability. The
regions implied in the cognitive task are represented by disks of
different colors. The populations of active neurons are not exactly at the
same
position across subjects
(top), and the across subjects mean signal in informative voxels  (bottom)
carries very weak information. Thus, it is clear that, in this case,
\emph{voxel-based} decoding approaches will perform poorly.
However, the mean of informative voxels within each region across
subjects (bottom) carries more information and should yield an accurate
inter-subject prediction.}
\label{Fig:intersubject}
\end{center}
\end{figure}

Given that an fMRI brain image has typically $10^4$ to $10^5$
voxels, it is perfectly reasonable to use
intermediate structures such
as parcels for reducing the dimensionality
of the data.
We also confirmed by different experiments that parcels are a good way
to tackle the spatial variability problem in inter-subjects studies. Thus
\emph{feature agglomeration} is an
accurate approach for the
challenging inter-subject generalization of
\emph{brain-reading} \cite{haxby2006,haynes2006}.
This can be explained by the fact that considering \emph{parcels} allows to
localize functional activity
across subjects and thus find a common support of neural codes of interest (see
Fig.~\ref{Fig:intersubject}). On the contrary, voxel-based methods suffer from 
the inter-subject spatial variability and their performances are
relatively lower.

Our approach entails the technical difficulty of optimizing the parcellation
with respect to the spatial organization of the information within the
image.
To break the combinatorial complexity of the problem, we have defined
a recursive parcellation of the volume using Ward algorithm, which
is furthermore constrained to yield spatially connected clusters. Note that it
is important to define the parcellation on the training database to
avoid data overfit.
The sets of possible volume parcellations is then reduced to a tree, and the 
problem reduces to finding the optimal cut of the tree.
We propose a \emph{supervised cut} approach that attempts to optimize the cut
with respect to the prediction task. Although finding an optimal
solution is infeasible, we adopt a  greedy
strategy that recursively finds the splits that most improve the
prediction score.  However, there is still no
guarantee that the optimal cut might be reached with this strategy.
Model selection is then performed a posteriori by considering the
best generalizing parcellation among the available models.
Additionally, our method is tractable on real data and runs in a very reasonable
of time (a few minutes without specific optimization).

In terms of \emph{prediction accuracy}, the proposed methods yield better
results
for the inter-subjects study on the different experiments, compared to
state of the art approaches (\emph{SVR}, \emph{Elastic net}, \emph{SVC} and
\emph{SMLR}).
The \emph{supervised cut} yields similar or higher prediction accuracy than the
\emph{unsupervised cut}. In the size prediction analysis,
the information is probably  coarser than in the object prediction analysis,
and thus the simple heuristic
of \emph{unsupervised cut} yields a good prediction accuracy.
Indeed, the unsupervised clustering still optimizes a cost function by
selecting the number of parcels that maximizes the prediction accuracy. Thus,
in simple prediction task such as the regression problem detailed
in this article, this approach allows
to extract almost all the relevant information. 
However, in the prediction of more fine-grained information, such as in the
classification task, the UC procedure does not provide a sufficient exploration
of the different parcellations, and does not extract all the relevant
information. Contrariwise, the SC approach explores relevant parcellations
using supervised information, and thus performs better than UC.

In terms of \emph{interpretability}, we have shown on simulations and real data
that this approach has the particular capability to highlight regions of
interest, while leaving uninformative regions unsegmented, and it can be viewed
as a multi-scale segmentation scheme \cite{michel2010}. The proposed scheme is
further useful
to locate contiguous predictive regions and to create interpretable
maps, and thus can be viewed as an intermediate approach between brain
mapping and inverse inference.
Moreover, compared to a state of the art approach for fine-grained decoding,
namely the searchlight, the proposed method yields similar maps, but
additionally, takes into account non-local information and yields only one
prediction score corresponding to  whole brain analysis.
From a neuroscientific point of view, the proposed approach retrieves well-known
results, \emph{i.e.} that differences between sizes (or between stimuli with
different spatial envelope in general) are most accurately represented in the
signals of early visual regions that have small and retinotopically laid-out
receptive fields.

More generally, this approach is not restricted to a given prediction function
and can be used with many different classification/regression methods.
Indeed, by restricting the search of the best subset of voxels to a tree pruning
problem, our algorithm allows us to guide the construction of the prediction
function in a low-dimensional representation of a high-dimensional dataset.
Moreover, this method is not restricted to brain images, and might be
used in any dataset where multi-scale structure is considered as
important (e.g. medical or satellite images).

In conclusion, this paper proposes a method for extracting information from
brain images, that builds relevant features by \emph{feature
agglomeration} rather than simple selection.
A particularly important property of this approach is its ability to
focus on relatively small but informative regions while leaving vast
but uninformative areas unsegmented.
Experimental results demonstrate that this algorithm performs well for
inter-subjects analysis where the accuracy of the prediction is tested on new
subjects. Indeed, the spatial averaging of the signal induced by the
parcellation appears as a powerful way to deal with inter-subject
variability.
\bibliographystyle{elsarticle-num}
\bibliography{sc}

\end{document}